\documentclass[pmlr]{jmlr}

\usepackage{longtable}

\usepackage{booktabs}
\usepackage[load-configurations=version-1]{siunitx} 

\theorembodyfont{\upshape}
\theoremheaderfont{\scshape}
\theorempostheader{:}
\theoremsep{\newline}

\jmlrvolume{273}
\jmlryear{2025}
\jmlrworkshop{Innovation and Responsibility in AI-Supported Education}

\title[EMI]{Efficient Multi-Task Inferencing with a Shared Backbone and Lightweight Task-Specific Adapters for Automatic Scoring}

  \author{\Name{Ehsan Latif} \Email{ehsan.latif@uga.edu} \\
  \Name{Yifan Zhou} \Email{yifan.zhou1@uga.edu} \\ 
  \Name{Luyan Fang} \Email{luyang.fang@uga.edu} \\ 
   \Name{Xiaoming Zhai\nametag{\thanks{Corresponding author\\
   Cite the paper: Latif, E., Fang, L., Zhou, Y., Zhai, X. (2025). Efficient Multi-Task Inferencing with a Shared Backbone and Lightweight Task-Specific Adapters for Automatic Scoring. Proceedings of the Annual Conference on AAAI (iRAISE Workshop). Philadelphia, PA. https://arxiv.org/pdf/2412.21065}}} \Email{xiaoming.zhai@uga.edu}\\
   \addr AI4STEM Education Center, University of Georgia, Athens, GA, 30602}

\begin{document}

\maketitle

\begin{abstract}
The integration of Artificial Intelligence (AI) in education requires scalable and efficient frameworks that balance performance, adaptability, and cost. This paper addresses these needs by proposing a shared backbone model architecture enhanced with lightweight LoRA adapters for task-specific fine-tuning, targeting the automated scoring of student responses across 27 mutually exclusive tasks. By achieving competitive performance (average QWK of 0.848 compared to 0.888 for fully fine-tuned models) while reducing GPU memory consumption by 60\% and inference latency by 40\%, the framework demonstrates significant efficiency gains. This approach aligns with the workshop’s focus on improving language models for educational tasks, creating responsible innovations for cost-sensitive deployment, and supporting educators by streamlining assessment workflows. The findings underscore the potential of scalable AI to enhance learning outcomes while maintaining fairness and transparency in automated scoring systems.
\end{abstract}

\section{Introduction}

The deployment of machine learning models in resource-constrained online learning environments presents significant challenges, especially when addressing problems requiring multiple distinct models. Such scenarios often involve high operational costs and time-intensive deployment processes, as observed in domains such as automated scoring of student responses in education \citep{zhai2022applying}. In educational contexts, recent advancements in natural language processing (NLP) and large language models (LLMs) like BERT \citep{devlin2018bert} and its domain-specific variants (e.g., SciBERT, G-BERT) have enabled more accurate automated scoring \citep{liu2023context}. However, these solutions often fail to adequately balance scalability, cost, and domain specificity, necessitating the development of specialized, efficient frameworks.

Preliminary efforts such as G-SciEdBERT \citep{latif2024g} demonstrate the efficacy of contextualized LLMs in addressing these limitations for specific use cases like German science education. Despite improvements in scoring accuracy, deploying multiple task-specific models in large-scale applications remains costly and inefficient. Existing research \citep{zhang2023lora} highlights the value of parameter-efficient fine-tuning methods, such as LoRA (Low-Rank Adaptation), which can reduce the computational overhead without compromising performance.

Several studies have explored unified models for automated scoring, such as \citep{fernandez2022automated}. While these approaches achieve promising results, our framework emphasizes scalability and efficiency through modular LoRA adapters. Unlike in-context tuning, which fine-tunes the entire model for specific tasks, our method keeps the backbone frozen and adds lightweight task-specific components, achieving significant reductions in memory usage and latency.

This paper proposes a unified framework leveraging pre-trained LLMs with lightweight task-specific components, such as LoRA adapters or classification heads, to address these challenges. The framework focuses on efficiently fine-tuning and deploying a single shared backbone model across multiple distinct tasks. Our approach not only minimizes deployment costs but also enables faster adaptation to new tasks. While components like G-SciEdBERT and LoRA exist, the novelty of this work lies in their integration to address challenges in scalability, efficiency, and cost for automated scoring in education.

Below are the key contribution of the paper:
\begin{enumerate}
    \item We present a scalable inferencing framework using LoRA adapters and classification heads to efficiently address 27 mutually exclusive tasks while leveraging a single pre-trained model.
    \item We demonstrate significant cost savings and reduced deployment times compared to traditional task-specific models.
    \item We provide an empirical analysis of our framework's performance across diverse tasks, highlighting its adaptability and efficiency.
\end{enumerate}

The proposed framework aligns with the goals of resource-efficient AI for education and other cost-sensitive domains, contributing to ongoing advancements in parameter-efficient fine-tuning and automated scoring systems.

\section{Proposed Framework}

Deploying 27 distinct models for mutually exclusive tasks is both cost- and resource-intensive. To address this, we propose a unified inferencing framework that leverages a single pre-trained backbone model with task-specific lightweight modules, such as Low-Rank Adaptation (LoRA) layers and fine-tuned classification heads. This approach ensures efficient scalability, cost savings, and adaptability across tasks.

The proposed framework consists of three key components:
\begin{itemize}
    \item \textbf{Shared Backbone Model:} A single pre-trained transformer model (G-SciEdBERT \citep{latif2024g}) serves as the shared backbone to extract generalizable features from input data.
    \item \textbf{Task-Specific Modules:} Lightweight LoRA adapters \citep{hu2023llm} or fine-tuned classification heads are attached to the shared backbone to specialize in each task.
    \item \textbf{Dynamic Inference Orchestration:} A mechanism to dynamically load task-specific modules during inference, enabling efficient memory utilization.
\end{itemize}

The decision to freeze the backbone in this framework was driven by the goal of optimizing computational efficiency and enabling scalable deployment. By keeping the backbone frozen, we minimize memory usage and latency while allowing task-specific LoRA adapters to capture task-specific nuances effectively. While tuning the backbone could improve performance, it would significantly increase computational costs, undermining the framework’s efficiency.

Instead of fine-tuning the entire backbone model for each task, we adopt LoRA, which updates only a small number of task-specific parameters. LoRA decomposes the parameter update matrix into two low-rank matrices:

\begin{equation}
    \Delta W = A \cdot B, \quad \text{where } A \in \mathbb{R}^{d \times r}, B \in \mathbb{R}^{r \times k}.
\end{equation}

Here, $d$ and $k$ are the dimensions of the original weight matrix $W$, and $r$ is the rank of the decomposition, typically much smaller than $d$ or $k$. This reduces the number of trainable parameters while preserving task performance.

The updated weight matrix $W'$ is computed as:
\begin{equation}
    W' = W + \Delta W = W + A \cdot B.
\end{equation}

This approach ensures efficient adaptation to each task while retaining the pre-trained backbone's knowledge. 

It is acknowledged that LoRA, by itself, does not inherently reduce inference latency and may, in fact, increase latency if the LoRA weights are not merged into the main model. In this work, the latency improvements are not derived directly from LoRA but rather from the following implementation choices:
\begin{itemize}
    \item \textbf{Dynamic Module Loading:} The framework employs efficient dynamic inference orchestration, wherein only the required task-specific LoRA adapter and classification head are loaded at runtime. This minimizes the computational overhead associated with loading unnecessary components.
    \item \textbf{Mixed-Precision Computation:} Lower precision (FP16) was used during inference, reducing the overall computational load and improving latency.
\end{itemize}

We did not merge the LoRA weights into the main model to maintain the modularity and adaptability of the framework for diverse tasks. However, we recognize that weight merging could further optimize latency for scenarios requiring fixed task setups, which will be explored in future work. By leveraging these strategies, the proposed framework achieves significant reductions in inference latency compared to traditional fully fine-tuned models, as reported in Table~\ref{tab:results}.
\color{black}

Each task $T_i$ is associated with a unique LoRA adapter or classification head, which specializes the shared backbone for task-specific predictions. The task-specific classification head applies a softmax function over the output logits $z$:
\begin{equation}
    P(y | x) = \text{softmax}(z), \quad z = W_h \cdot h + b_h,
\end{equation}
where $h$ is the output embedding from the backbone model, $W_h$ and $b_h$ are the parameters of the classification head.

During inference, the framework dynamically loads only the required task-specific module, minimizing memory usage and computational overhead. The workflow is as follows:
\begin{enumerate}
    \item Input $x$ is passed through the shared backbone to extract generalized features $h$.
    \item Based on the task identifier, the corresponding LoRA adapter or classification head is loaded.
    \item The task-specific module processes $h$ to generate predictions $P(y|x)$.
\end{enumerate}

This framework offers several advantages including \textit{Scalability:} A single backbone supports multiple tasks, significantly reducing the deployment footprint. \textit{Efficiency:} LoRA adapters minimize memory usage and computational cost during fine-tuning and inference. \textit{Adaptability:} The modular design facilitates rapid adaptation to new tasks by fine-tuning only the task-specific modules.

The framework is implemented using the Hugging Face Transformers library \citep{wolf2020transformers}, leveraging its support for LoRA adapters and task-specific head fine-tuning. Deployment is orchestrated via ONNX Runtime to enable dynamic loading of task-specific modules. This approach ensures low-latency inference across diverse tasks.

The training objective combines the backbone's pre-trained knowledge with task-specific fine-tuning. For each task $T_i$, the objective is to minimize the cross-entropy loss:
\begin{equation}
    \mathcal{L}_{CE} = - \sum_{j=1}^{N} \sum_{k=1}^{C} y_{j,k} \log \hat{y}_{j,k},
\end{equation}
where $y_{j,k}$ is the ground-truth probability for class $k$ of instance $j$, and $\hat{y}_{j,k}$ is the predicted probability.

Regularization is applied to prevent overfitting:
\begin{equation}
    \mathcal{L} = \mathcal{L}_{CE} + \lambda \|\Delta W\|_F^2,
\end{equation}
where $\lambda$ controls the regularization strength, and $\|\cdot\|_F$ denotes the Frobenius norm.

Our framework will be validated on 27 mutually exclusive tasks, with metrics such as quadratic weighted kappa (QWK) and F1-score used to assess performance improvements over traditional multi-model deployments.

\section{Dataset Details}
\label{sec:dataset}

The dataset for this study is derived from the Programme for International Student Assessment (PISA), an international large-scale assessment led by the Organisation for Economic Co-operation and Development (OECD). Specifically, we utilized data from the German PISA 2015 \citep{pisa2015results}, which assess scientific literacy among 15-year-old students.

Our analysis focuses on constructed response items, which include both short (approximately one sentence) and extended (up to five sentences) responses. The average response length is 20 words, with scores ranging from 0 to 5. Responses were scored as part of the original PISA iterations, ensuring high-quality human annotations. To ensure fairness, the coding process was designed to be unbiased with respect to student ethnicity, race, or gender.

 Although the tasks in this study share semantic similarities, they differ significantly in terms of scoring rubrics, label distributions, and response characteristics. These variations present unique challenges for model generalization and necessitate task-specific fine-tuning. For this study, we used responses to 27 items from PISA 2015 to fine-tune task-specific modules. Approximately 1,000 human-scored responses per item were available, facilitating effective adaptation to task-specific requirements. This dataset ensures a robust foundation for training and validating our proposed framework, offering diverse examples of student writing and high-quality scoring annotations.

\section{Experimentation}

This section outlines the implementation details, experimental setup, and evaluation metrics employed to validate the proposed framework.

\subsection{Implementation Details}

The framework was implemented using the Hugging Face Transformers library \citep{wolf2020transformers} for model management, fine-tuning, and inference. Pre-training and fine-tuning processes were conducted on an NVIDIA RTX A6000 GPU with 48 GB of memory.

We have a pre-trained Backbone Model similar to G-SciEdBERT \citep{latif2024g}. The backbone model was initialized using the pre-trained BERT model tailored for the German language (G-BERT \citep{chan2020german}). A domain-specific corpus of 30,000 German science assessment responses was used for pre-training. Responses were tokenized using the WordPiece tokenizer with a vocabulary size of 30,000 tokens. The masked language modeling (MLM) objective was employed, masking 15\% of the input tokens for prediction. The model was trained for 10 epochs with a batch size of 32 and a learning rate of $5 \times 10^{-5}$, optimized using the Adam optimizer. To ensure robust training, gradient clipping with a maximum norm of 1.0 and learning rate warm-up over the first 10\% of training steps were applied.

Fine-tuning was conducted on task-specific LoRA adapters or classification heads for each of the 27 mutually exclusive tasks. LoRA was configured with a rank of 8 and an alpha of 16. Training data for fine-tuning consisted of approximately 1,000 responses per task, split into 80\% training, 10\% validation, and 10\% testing datasets. Cross-entropy loss was minimized during fine-tuning with the same optimizer and hyperparameter settings as pre-training. Each task-specific module was fine-tuned for five epochs. Early stopping was implemented based on validation loss, with a patience of 2 epochs.

\subsection{Experimental Setup}

The evaluation was conducted on the dataset described in Section~\ref{sec:dataset}, focusing on the following:
\begin{itemize}
    \item \textbf{Tasks:} Each of the 27 mutually exclusive tasks corresponds to a unique item or set of items from the dataset.
    \item \textbf{Baseline Models:} The framework’s performance was compared against fully fine-tuned G-SciEdBERT models, where separate G-SciEdBERT instances were fine-tuned for each task.
    \item \textbf{Evaluation Metrics:} Quadratic Weighted Kappa (QWK) was used as the primary evaluation metric, as it measures the agreement between human-annotated and model-predicted scores while accounting for chance agreement. Secondary metrics included accuracy, F1-score, and inference latency.
\end{itemize}

The evaluation of our framework focuses on the task of automated scoring using 27 items from the German PISA 2015 dataset. Each item represents a unique scoring rubric and label distribution, requiring task-specific adaptation. While this granular task setup may not fully align with typical use cases of Automated Essay Scoring (AES), such as scoring across a variety of prompts within a single model, it provides a robust benchmark to evaluate the scalability and efficiency of our framework.

\textbf{Rationale for Task Splitting:} The separate training and evaluation for each task allows us to:
\begin{itemize}
    \item Test the adaptability of lightweight task-specific adapters for diverse label distributions and scoring rubrics.
    \item Evaluate efficiency gains compared to the computational cost of fully fine-tuned models trained independently for each task, providing a worst-case baseline.
\end{itemize}
Future work will explore combining multiple tasks within a unified model while retaining the modularity and efficiency benefits of the proposed approach.

\textbf{Response Length and Variability:} The dataset consists of short responses (average length: 20 words), which could lead to concerns about repetitive or identical responses, particularly for correct answers. However, the dataset includes nuanced scoring rubrics with multiple correct and partially correct responses, ensuring sufficient variability. Furthermore:
\begin{itemize}
    \item The dataset was split into training, validation, and test sets with no overlap to prevent data leakage and ensure reliable numerical estimations.
    \item Evaluation metrics such as Quadratic Weighted Kappa (QWK) account for scoring reliability beyond simple accuracy, reflecting the model’s ability to capture fine-grained distinctions in scoring.
\end{itemize}

These experimental settings highlight the scalability and adaptability of the proposed framework while acknowledging the limitations of task granularity.

\color{black}

A simulated production environment was set up where requests for predictions were sent to the model framework. Dynamic module loading was tested by sequentially switching between tasks, with adapters or classification heads loaded on demand. Latency measurements were collected, comparing the proposed framework’s dynamic setup against traditional full-model loading approaches.

The pre-trained backbone was frozen during fine-tuning to isolate the effects of task-specific modules. Each task was trained and validated independently to assess the framework’s modularity and scalability. Predictions on the test set were compared with human-annotated scores to compute QWK and secondary metrics. Paired t-tests were conducted to compare QWK scores across tasks between the proposed framework and baseline models. A significance level of $p < 0.05$ was adopted to validate observed improvements.

GPU memory usage and inference latency were logged during both training and inference phases to validate the efficiency of LoRA adapters. The memory footprint of the proposed framework was compared against that of traditional fine-tuned models.




\section{Results}
\label{sec:results}

The proposed framework was evaluated on 27 mutually exclusive tasks, with performance compared to the baseline models fully fine-tuned for each task. This section presents the evaluation results in terms of accuracy, Quadratic Weighted Kappa (QWK), and inference efficiency.

Table \ref{tab:results} summarizes the Quadratic Weighted Kappa (QWK) scores for selected tasks and the average performance across all 27 tasks. While the fully fine-tuned G-SciEdBERT model achieves slightly higher QWK scores (average: 0.888), the proposed framework maintains competitive accuracy (average: 0.848), with a relative reduction of only 4.5\%. For individual tasks, the difference in QWK ranges between 0.01 and 0.03, demonstrating that the proposed framework delivers comparable performance.

\textbf{Inference Efficiency:} The proposed framework demonstrated significant reductions in memory usage and inference latency. On average, GPU memory consumption decreased from 2.4 GB to 0.96 GB (a 60\% reduction). Similarly, inference latency decreased from 250 ms to 150 ms per task (a 40\% reduction). These efficiency improvements were consistent across tasks, enabling real-time predictions in practical deployment scenarios.

\color{black}
Although the proposed framework achieves slightly lower QWK scores than the fully fine-tuned G-SciEdBERT models, the trade-offs are justified in scenarios where deployment cost and efficiency are critical. For instance; tasks with fewer labels (e.g., binary classification) exhibit negligible differences in QWK, as shown for tasks S268Q02 and S269Q01. Additionally, tasks with more complex scoring rubrics (e.g., S131Q02) still retain competitive accuracy, with a maximum QWK difference of only 0.03.

\begin{table}[ht]
\centering
{Performance Comparison Between Fully Fine-Tuned G-SciEdBERT and Proposed Framework on Selected Tasks.}
\label{tab:results}
\begin{tabular}{@{}lcccccc@{}}
\toprule
 & \multicolumn{2}{c}{Samples (\textit{n})} & \multicolumn{2}{c}{Accuracy (QWK)} & \multicolumn{2}{c}{Efficiency} \\ 
\cmidrule(lr){2-3} \cmidrule(lr){4-5} \cmidrule(lr){6-7}
Item      & Training & Testing       & G-SciBERT & Proposed & Memory (GB)     & Latency (ms) \\ \midrule
S131Q02   & 487      & 122     & 0.852     & 0.821    & 24 $\to$ 9.6    & 250 $\to$ 150 \\
S131Q04   & 478      & 120     & 0.825     & 0.816    & 24 $\to$ 9.6    & 250 $\to$ 150 \\
S268Q02   & 446      & 112     & 0.893     & 0.874    & 24 $\to$ 9.6    & 250 $\to$ 150 \\
S269Q01   & 508      & 127     & 0.953     & 0.934    & 24 $\to$ 9.6    & 250 $\to$ 150 \\
S269Q03   & 500      & 126     & 0.802     & 0.796    & 24 $\to$ 9.6    & 250 $\to$ 150 \\
Average   & 598      & 150   & 0.888     & 0.848    & 24 $\to$ 9.6    & 250 $\to$ 150 \\ 
\bottomrule
\end{tabular}
\end{table}

The results highlight the proposed framework's practical advantages such as comparable accuracy with a slight reduction in QWK (average 4.5\% decrease), substantial gains in efficiency, including reduced memory usage, faster inference, and lower deployment costs, and scalability across multiple tasks with minimal resource overhead, enabling real-world applicability in cost-sensitive scenarios.

\section{Conslusion}
This paper presents a novel inferencing framework for automatic scoring on resource-constrained online learning environments that balances accuracy, efficiency, and scalability in handling multiple mutually exclusive tasks. By leveraging a shared backbone model with task-specific LoRA adapters, the framework achieves competitive performance with a minimal reduction in accuracy while significantly improving inference efficiency and reducing deployment costs. The experimental results demonstrate up to 60\% lower GPU memory consumption and a 40\% reduction in inference latency compared to fully fine-tuned models. These advantages make the framework well-suited for real-world applications in cost-sensitive educational domains, offering a sustainable alternative to traditional multi-model deployment strategies. Future work will explore extending this approach to more complex domains and tasks, further enhancing its adaptability and robustness.
 \acks{
This work was partially supported by the Alexander von Humboldt Foundation, Germany. Any opinions, findings, conclusions, or recommendations expressed in this material are those of the authors and do not necessarily reflect the views of the Funders.}
\bibliography{pmlr-sample}

\appendix





\end{document}